%% file: main.tex
\pdfoutput=1

\documentclass[11pt]{article}

\usepackage[]{acl}

\usepackage{times}
\usepackage{latexsym}
\usepackage{multirow}

\usepackage[T1]{fontenc}

\usepackage[utf8]{inputenc}

\usepackage{microtype}

\usepackage{inconsolata}
\usepackage{multirow}

\usepackage{enumitem}
\usepackage{graphicx}
\usepackage{amsmath}

\title{Consistent Joint Decision-Making with Heterogeneous Learning Models}

\author{Hossein Rajaby Faghihi \\
  Michigan State University \\
  \texttt{rajabyfa@msu.edu} \\\And
  Parisa Kordjamshidi \\
  Michigan State University \\
  \texttt{kordjams@msu.edu}}

\begin{document}
\maketitle
\begin{abstract}

This paper introduces a novel decision-making framework that promotes consistency among decisions made by diverse models while utilizing external knowledge. Leveraging the Integer Linear Programming~(ILP) framework, we map predictions from various models into globally normalized and comparable values by incorporating information about decisions' prior probability, confidence~(uncertainty), and the models' expected accuracy.
Our empirical study demonstrates the superiority of our approach over conventional baselines on multiple datasets.
\end{abstract}

\section{Introduction}
The rapid advance of AI has led to the widespread use of neural networks in tackling complex tasks that involve multiple output decisions, which may be derived from various models~\cite{liu2022autoregressive,wang2022deepstruct}. However, in many cases, these decisions are interrelated and must conform to specific constraints. For example, to comprehend procedural text, multiple neural models collaborate to establish temporal relationships between actions, reveal semantic relations, and discern entity properties like location and temperature~\cite{faghihi2023role, bosselut2017simulating, jiang2023transferring}. Each model exhibits distinct decision characteristics, output sizes, uncertainty levels, and varying excepted accuracy levels. Resolving inconsistencies and aligning these diverse neural decisions is crucial for a comprehensive understanding of the underlying process.

In many instances, raw model outputs lack usability without enforcing consistency. In tasks like hierarchical image classification, with independent models for each hierarchy level, outputs should adhere to the known hierarchical relationships. For example, the combination ``Plant, Chair, Armchair'' lacks validity and requires post-processing for downstream applications. A similar requirement extends to generative models in text summarization~\cite{lu2021neurologic} and image captioning~\cite{anderson2017guided}.
Prior studies have proposed techniques for handling inconsistencies in correlated decisions during both inference~\cite{freitag2017beam,scholak2021picard,dahlmeier2012beam,chang2012structured,guo2021inference} and training~\cite{hu2016harnessing,nandwani2019primal,xu2018semantic} of neural models. This paper focuses on resolving these inconsistencies at inference, where the goal is to ensure that outputs align with task constraints while preserving or enhancing the original model performance without training.

\input{image/overview}
In addressing decision inconsistencies, Integer Linear Programming (ILP)~\cite{roth2005integer} stands out as a robust approach. ILP is a global optimization framework that seeks to find the best assignments to variables while meeting specified constraints. It is known for its efficiency and capability to produce globally optimal solutions, distinguishing it from alternatives like beam search. The ILP formulation is as follows:
\vspace{-2mm}
\begin{equation}
\vspace{-2mm}
\label{formula:ilp}
\begin{aligned}
\operatorname{Objective} &: \operatorname{Maximize}  \quad {P}^\top y \\
            & \mbox{subject to} \quad \mathcal{C}\left(y\right)\le0,
\end{aligned}
\end{equation}
where constraints are denoted by $\mathcal{C}\left(\cdot\right) \le 0$, decision variables are denoted by $y\in\mathcal{R}^n$, and the vector containing the local weights of variables (i.e coefficients of the output variables in the objective function) are denoted by $P$. In order to apply ILP to resolve conflicts from decisions of neural models, prior work~\cite{rizzolo2016integer, punyakanok2004semantic,ning2018joint,inference-ijcai2020-382,SemanticWeb} has defined $P$ to be the vector of raw probabilities of local decisions, $P = [p^1, ..., p^n]$, where $p^i$ corresponds to the probability generated from a certain model for the $i$th decision variable~($y_i$). The global inference is modeled to maximize the combination of probabilities subject to constraints. 
Although the constraints can take any form of equality or inequality applied on combinations of $y$ variables, here, we focus on logical constraints. We utilize the mapping of logical constraints to $y$ equations introduced in the DomiKnowS~\cite{faghihi2021domiknows} framework. For instance, in order to map the mutual exclusivity constraint in a multi-class classification task $C$ with $N$ possible outputs, where decision variables over a single input are expressed by $\{(Y_C^1, P_C^1), (Y_C^2, P_C^2), ..., (Y_C^N, P_C^N)\}$, the constraint is expressed as $\sum_{i=1}^N Y_{C}^i = 1$. 
Ignoring other variables and constraints, the optimization problem becomes, 
\begin{equation}
\operatorname{Maximize} \quad \sum_{i=1}^N Y_{C}^i P_C^i \quad \text{s.t.} \quad \sum_{i=1}^N Y_{C}^i = 1.
\end{equation}
In this simplification, since the problem is to find $Y_C^i$ values in integer space, the best solution sets the $Y_C^i$ value to 1 for the $i$th element that has the largest $P_C^i$. The rest of the values are set to zero.
Previous use of ILP has proven effective in ensuring decision consistency in certain cases~\cite{faghihi2023gluecons} but did not address model heterogeneity. This problem becomes more dominant in scenarios where output probabilities come from independent models, making them less directly comparable. To address this limitation, we extend the ILP formulation beyond just considering the raw model probabilities. Instead, we map these raw scores into globally comparable values, facilitating a more balanced global optimization.
We achieve this by incorporating additional information, such as decision confidence, expected model accuracy, and estimated prior probabilities. 
While previous studies have explored the integration of uncertainty in modeling the training objective~\cite{xiao2019quantifying,gal2016dropout,zhu2017deep}, our work represents a novel effort in systematically incorporating multiple factors of this nature into the inference process for interrelated decisions to leverage external knowledge effectively.

The methods proposed in this paper are now publicly available and have been properly integrated into the ILP inference pipeline of the DomiKnowS framework\footnote{\url{https://github.com/HLR/DomiKnowS}}. 

\section{Method}
Figure \ref{fig:overview} shows an overview of our general framework and its components. 
Our objective is to devise an improved scoring system, generating new local variable weights~(importance) $W$ in the ILP formulation. Thus, we modify the original objective function as follows:
\vspace{-2mm}
\begin{equation}
\vspace{-2.2mm}
\label{formula:main}
\begin{aligned}
\operatorname{Maximize} \quad {W}^\top y,
\end{aligned}
\end{equation}
where $W = [w^1, ..., w^n]$. To determine the new weights, we aim to find the scoring function $G$, which normalizes the local predictions of each model and maps them into globally comparable values. For each model $m$ with multi-class decisions, we denote the output probabilities after applying a SoftMax layer as $P_m \subset P$. The scoring function $G$ transforms these raw probabilities into new weights $W_m \subset W$ to indicate the importance of the variables within the ILP objective, i.e., $W_m = G(P_m, m)$. 
This section explores different options for the function $G$ and provides an intuitive understanding of their rationale. 


\subsection{Prior Probability~(Output Size)}
To facilitate fair comparison among decisions with varying output sizes, we consider a normalization factor based on prior probabilities. For an $N$-class output, the prior probability for each label is $\frac{1}{N}$~(assuming uniform distribution). This implies an inherent disadvantage for decisions made in larger output spaces. Thus, we normalize the raw probabilities by dividing them by the inverse of their respective priors and define $G(P_m, m) = P_m \times N$. 
\subsection{Entropy and Confidence}
The outputs generated from models often exhibit varying levels of confidence. While raw probabilities alone may adequately indicate the model's confidence in individual Boolean decisions, a more sophisticated approach is required for assessing the models' confidence in multi-classification.
We propose incorporating the entropy of the label distribution as an additional factor to assess the model's decision-making confidence. As lower entropy corresponds to higher confidence, we use the reverse of the entropy, normalized by the output size $N$, as a factor in forming the decision weight function $G(P_m, m) = P_m * (\frac{N}{Entropy(P_m)})$.

\subsection{Expected Models' Accuracy}
Assigning higher weights to the probabilities generated by more accurate models aligns the optimal solution with the overall underlying models' performance. This approach mitigates the influence of poor-quality decisions, which can negatively impact others in the global setting. We define the decision weight function $G$ as $G(P_m, m) = P_m * Acc_{m}$, where $Acc_{m}$ represents the accuracy of the corresponding model, measured in isolation. To mimic the real-world settings where test labels are not available during inference, we utilize the models' accuracies on a probe/dev set.




\section{Empirical Study}
We assess the impact of integrating proposed factors into the ILP formulation on a series of structured prediction tasks. Our approach is particularly suited for hierarchical structures encompassing multiple classes at different granularity levels, such as classical hierarchical classification problems. Additionally, we are the first to investigate the influence of enforcing global consistency on the procedural reasoning task, a complex real-world problem.
To implement our method, we rely on the DomiKnowS framework~\cite{rajaby-faghihi-etal-2021-domiknows, faghihi2023gluecons}, offering a versatile platform that enables implementing and evaluating techniques to leverage external logical knowledge with minimal effort on structured output prediction tasks.

\subsection{Metrics and Evaluation}
We compare our method against two inference-time approaches: sequential decoding and basic ILP~(ILP without our refinement). In contrast to ILP, sequential decoding, which relies on expert-designed rules or programs to enforce consistency, is unique to each dataset.
In addition to conventional metrics~(e.g., accuracy/F1), we include measurements that evaluate changes applied by the inference techniques: (1) total changes~(\textbf{C}), (2) the percentage of incorrect-to-correct changes~(\textbf{+C}), (3) the percentage of correct-to-incorrect changes~(\textbf{-C}). We further evaluate all the baselines and inference methods on (1) the percentage of decisions satisfying task constraints and (2) Set Correctness, the percentage of correct sets of interrelated decisions~(i.e., predictions of all levels in the hierarchy must be correct for an image).
More details are in Appendix \ref{sec:metrics_appendix}.

\subsection{Tasks}
We choose a set of tasks that contain multiple decisions with differences in output size, complexity, and availability of training data while still correlated in the same task. Our primary objective is to demonstrate that the new formulation for ILP can better align decisions in a heterogeneous space, thereby enabling better utilization of constraints to draw more accurate answers from models during inference. To achieve this, we have not necessarily selected state-of-the-art models as our baselines for all tasks. This is because we need to provide baselines where the model is not already completely aligned with the constraints, and the decisions can still benefit from applying constraints during inference. We showcase our method on both toy tasks and real-world tasks.
\subsubsection{Procedural Reasoning}
\noindent\textbf{Task: } 
Procedural reasoning tasks entail the tracking of entities within a narrative. Following ~\citet{faghihi2021timestamped}, we formulate this task as  Question-Answering~(QA). Two key questions are addressed for each entity $e$ and step $i$: (1) \textit{Where is $e$ located in step $i$?} and (2) \textit{What action is performed on $e$ at step $i$?}. The decision output of this task exhibits heterogeneity, encompassing a diverse range of possible actions~(limited multi-class) and varied locations derived from contextual information~(spans). The task constraints establish relationships between action and location decisions as well as among action decisions at different steps. For instance, the sequence of `Destroy, Move' represents an invalid assignment for action predictions at steps $i$ and $i+1$.

\noindent\textbf{Dataset: }
We utilize the \textbf{Propara} dataset~\cite{dalvi2018tracking}, a small dataset focusing on natural events. This dataset provides annotations for involved entities and their corresponding location changes. The label set is further expanded to include information on actions, which can be inferred from the sequence of locations. 

\noindent\textbf{Baseline: }
We employ a modified version of the MeeT architecture~\cite{singh-etal-2023-entity} as our baseline for this task. The MeeT model is designed to ask the two aforementioned questions at each step and employs a generative model~(T5-large) to answer those questions.
The \textbf{Sequential Decoding} baseline resolves action inconsistencies in a sequential stepwise manner~(first to last), followed by the selection of locations accordingly.
Additional information can be found in Appendix~\ref{sec:datasets}

\subsubsection{Hierarchical Classification}
\noindent\textbf{Task: } 
This task involves creating a hierarchical structure of parent-child relationships by classifying inputs into various categories at distinct levels of granularity.

\noindent\textbf{Datasets: }
We employ three different datasets. (1) A subset of the Flickr dataset~\cite{young2014image} with two hierarchical levels for the classification of images with types of \textit{Animal, Flower, and Food}, (2) 20News dataset for text classification, where the label set is divided into two levels, and (3) The OK-VQA benchmark~\cite{marino2019ok}, a subset of the COCO dataset~\cite{lin2014microsoft}. In OK-VQA, the hierarchical relations between labels are established into four levels based on ConceptNet triplets and the dataset's knowledge base.

\noindent\textbf{Baselines: } 
ResNet~\cite{he2016deep} and BERT~\cite{kenton2019bert} are used to obtain representations for the image and text modalities, respectively. Linear classification layers are applied to convert obtained representations into decisions. The \textbf{Sequential Decoding} is top-down, bottom-up, and a two-stage (1) top-down on `None' values and (2) bottom-up on labels for Animal/Flower/Food, 20 News, and VQA tasks, respectively. 
More information is available in Appendix \ref{sec:datasets}.

\subsection{Results}
Tables \ref{tab:animal}, \ref{tab:vqa}, and \ref{tab:propara} display results for \textit{Animal/Flower/Food}, \textit{Ok-VQA}, and \textit{Propara} datasets. Due to space constraints, results for the \textit{20News} dataset are in Appendix \ref{sec:appendix_20news}. For close results, we use multiple seeds to validate reliability.
Across experiments, the basic ILP technique favors decisions in smaller output spaces due to higher probability magnitudes~(e.g., more changes in Actions than Locations in Table \ref{tab:propara}).
Our new proposed variations can effectively mitigate this problem and perform a more balanced optimization.

\noindent\textbf{Animal/Flower/Food: } The sequential decoding establishes that the enforcement of the decisions originating from a model with better accuracy and with a smaller output size~(Level 1) on other decisions may even have a negative impact on them~(Level 2). In such scenarios, the inclusion of \textit{Expected Accuracy} favors dominant decisions and adversely affects performance. However, the inclusion of \textit{Prior Probability} proves effective in achieving a balanced comparison among decisions. In this task, despite the basic ILP formulation being detrimental, some of the new variations can even surpass the original baseline performance.
\input{tables/animal}
\input{tables/vqa}

\noindent\textbf{Ok-VQA: } The baseline exhibits lower accuracy in lower-level decisions with smaller output sizes. When applying the basic ILP method under these circumstances, a significant decline in results is observed, even below that of sequential decoding. However, incorporating any of our proposed factors leads to substantial improvements compared to the basic ILP formulation~(over $4\%$ improvement) and can surpass the performance of sequential decoding. Particularly, combining \textit{Entropy} and \textit{Prior Probability} achieves the best performance. Notably, although the baseline model has higher overall performance, its inconsistent outputs are unreliable for determining the object label~(see Table \ref{tab:new_metrics}).  
\input{tables/propara}
\input{tables/new_metrics}

\noindent\textbf{Propara: } This is an example of a real-world task that involves hundreds of constraints and thousands of variables when combining decisions across entities and steps. Once again, basic ILP and \textit{Expected Accuracy} factor prioritize decisions from the smaller output size~(Actions). However, the \textit{Prior probability} factor enables a more comparable space for resolving inconsistencies. Notably, the higher baseline performance is attributed to inconsistencies and cannot be used when reasoning about the process~(See Table \ref{tab:new_metrics}).

\noindent\textbf{Constraints: } Table \ref{tab:new_metrics} presents the results of satisfaction and set correctness metrics across various datasets. It is evident that our newly proposed method significantly outperforms the baseline in both of these metrics. Notably, the degree of improvement in set correctness is more pronounced when the initial consistency of the baseline is lower. This observation underscores the substantial significance of our proposed technique in ensuring the practical utility of model decisions in downstream applications by substantially increasing the proportion of correct interrelated decision sets.
Furthermore, in comparison to sequential decoding, our proposed solutions demonstrate even greater performance enhancements, particularly in scenarios where the task complexity is higher, and global inference can exert its maximum effectiveness.

\section{Conclusion}
This paper introduced an approach for taking into account the uncertainty and confidence measures, including the decisions' prior probability, entropy, and expected accuracy, alongside raw probabilities when making globally consistent decisions based on diverse models. Through experiments on four datasets, we demonstrated the effectiveness of incorporating our idea within the ILP formulation. This contribution presents a high potential in advancing large models by integrating them into a unified decision-making framework for conducting complex tasks requiring interrelated decisions.

\section*{Limitations}
Our implementation of Integer Linear Programming~(ILP) is based on the DomiKnowS framework, which relies on the Gurobi optimization engine~\cite{gurobi}. The availability of the Gurobi optimization engine in its free version is limited, which may pose constraints on the replication of our ILP-based approach for procedural reasoning experiments. However, the free academic license for Gurobi ensures the necessary access to execute all the tasks modeled in this paper.
It is important to note that while our experiments and discussions demonstrate the effectiveness of our proposed approach in addressing challenges encountered with conventional ILP utilization, it is not guaranteed to consistently yield improved performance in scenarios where the decision space of variables is already comparable or consists solely of boolean decisions. These limitations highlight the need for careful consideration and evaluation of the specific problem domain and characteristics when applying our approach or considering alternative methodologies.

\section*{Acknowledgements}
This project is supported by the National Science Foundation (NSF) CAREER award 2028626 and partially supported by the Office of Naval Research
(ONR) grant N00014-20-1-2005. Any opinions,
findings, and conclusions or recommendations expressed in this material are those of the authors and do not necessarily reflect the views of the National
Science Foundation nor the Office of Naval Research.
\bibliography{custom, previous}
\bibliographystyle{acl_natbib}

\appendix
\input{appendix}

\end{document}

%% file: image/overview.tex
\begin{figure*}
    \centering
    \includegraphics[width=\linewidth]{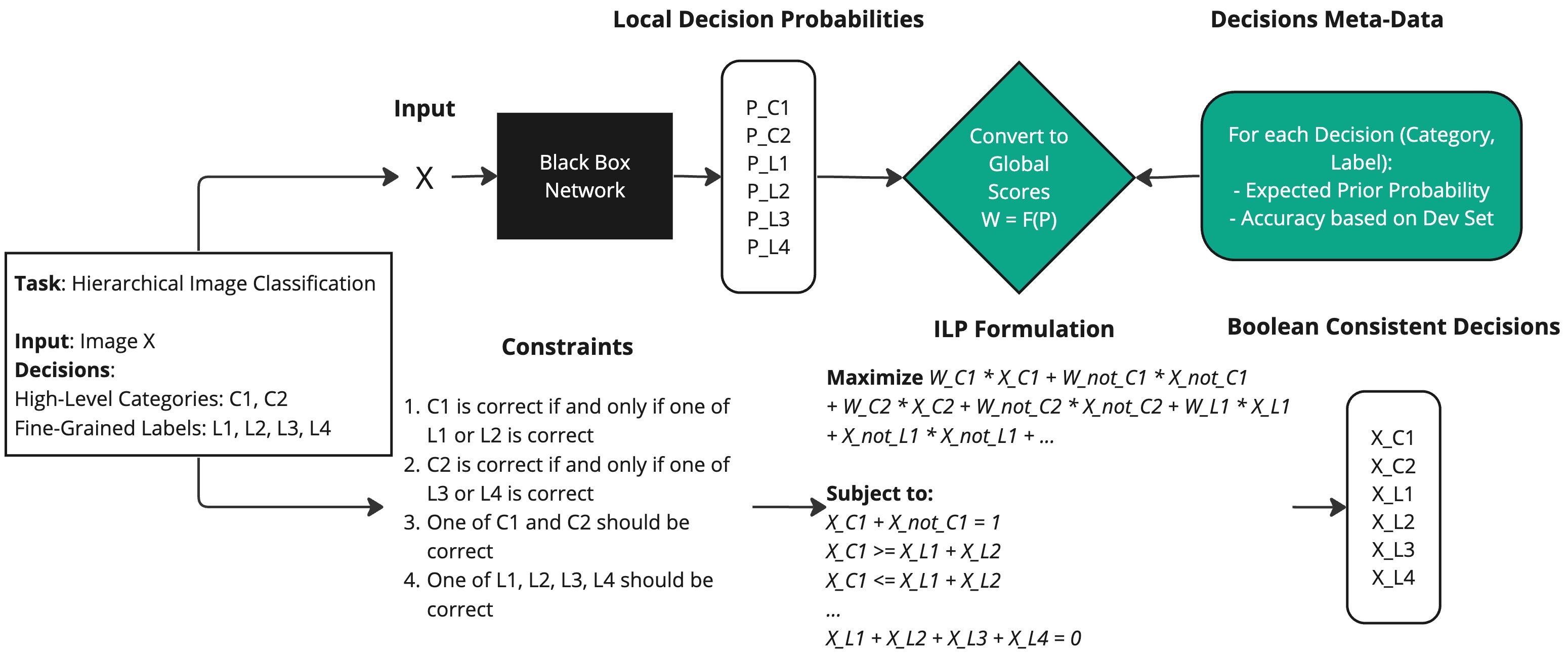}
    \caption{An overview of the proposed solution to maintain consistency between model decisions during inference via ILP optimization. The task used as an example here is the Hierarchical Image Classification task with two levels. The Green blocks represent additional components that have been added to the pipeline in this paper to guarantee the global comparability of model-generated probabilities.}
    \label{fig:overview}
\end{figure*}

%% file: tables/animal.tex
\begin{table}[ht]
\scriptsize
\setlength{\tabcolsep}{2.5pt}
\centering
\begin{tabular}{|c|cccc|ccccc|}
\hline
\multirow{2}{*}{Model} & \multicolumn{4}{c|}{Level 1 (3)}                   & \multicolumn{4}{c}{Level 2 (15)}                                         & Average        \\ \cline{2-10} 
                       & Acc            & C  & + C         & - C            & Acc            & C  & + C             & \multicolumn{1}{c|}{- C}         & Acc            \\ \hline
Baseline               & 86.12          & -  & -           & -              & \textbf{54.85} & -  & -               & \multicolumn{1}{c|}{-}           & 70.48          \\ \hline
Sequential             & 86.12          & -  & -           & -              & 54.39          & 32 & \textbf{15.625} & \multicolumn{1}{c|}{37.5}        & 70.25          \\ \hline
ILP                    & 86.07          & 16 & 43.75       & 43.75          & 54.43          & 16 & 12.5            & \multicolumn{1}{c|}{37.5}        & 70.25          \\
+ Acc                  & 86.14          & 3  & 33.33       & \textbf{33.33} & 54.41          & 29 & 13.79           & \multicolumn{1}{c|}{37.93}       & 70.27          \\
+ Prior                & 86.30          & 24 & 50          & 41.67          & 54.78          & 8  & 12.5            & \multicolumn{1}{c|}{\textbf{25}} & 70.54          \\
+ Ent + Acc            & 86.09          & 12 & 33.33       & 50             & 54.41          & 20 & 10              & \multicolumn{1}{c|}{40}          & 70.25          \\
+ Ent + Prior          & \textbf{86.42} & 25 & \textbf{52} & 40             & \textbf{54.82} & 7  & 14.29           & \multicolumn{1}{c|}{28.57}       & \textbf{70.62} \\
+ All                  & 86.17          & 16 & 43.75       & 43.75          & 54.50          & 16 & 12.5            & \multicolumn{1}{c|}{37.5}        & 70.33          \\ \hline
\end{tabular}
\caption{Results on \textit{Animal/Flower/Food} dataset on four random seeds. Reported values are the average scores of runs with close variances for all techniques{~\scriptsize(Level1: $\pm$1.6 and Level2: $\pm$0.5)}. \textbf{C} values are derived from the best run.
$n$ in \textbf{Level~($n$)} denotes the number of output space classes. \textbf{Prior:} Prior Probability, and \textbf{Ent:} Entropy.
}
\label{tab:animal}
\end{table}

%% file: tables/vqa.tex
\begin{table}[ht]
\centering
\footnotesize
\setlength{\tabcolsep}{3pt}
\begin{tabular}{|c|c|c|c|c|c|}
\hline
Model       & \begin{tabular}[c]{@{}c@{}}Level 1 \\ (274)\end{tabular} & \begin{tabular}[c]{@{}c@{}}Level 2 \\ (158)\end{tabular} & \begin{tabular}[c]{@{}c@{}}Level 3 \\ (63)\end{tabular} & \begin{tabular}[c]{@{}c@{}}Level 4 \\ (8)\end{tabular} & Average        \\ \hline

Baseline    & 56.73          & 54.45          & 43.43          & 17.68          & \textbf{54.64} \\ \hline
Sequential  & 55.81          & 53.17          & 43.44          & 24.18          & 53.72          \\ \hline
ILP         & 52.38          & 46.33          & \textbf{49.66} & \textbf{28.43} & 50.17          \\
+ Acc       & 55.65          & \textbf{54.67} & 48.15          & 23.73          & 54.23          \\
+ Prior       & 56.35          & 53.36          & 48.11          & 23.86          & 54.54          \\
+ Ent + Acc & 56.43          & 53.25          & 48.1           & 24.02          & 54.56          \\
+ Ent + Prior & 56.79          & 52.93          & 47.53          & 23.75          & \textbf{54.61} \\
+ All       & \textbf{56.84} & 52.66          & 46.98          & 22.63          & 54.5           \\ \hline
\end{tabular}
\caption{The results on the Ok-VQA dataset. The values represent the F1 measure. Levels 2, 3, and 4 contain `None' labels. The low F1 measure of lower levels is due to a huge number of False Positives. 
}
\label{tab:vqa}
\end{table}

%% file: tables/propara.tex
\begin{table}[ht]
\scriptsize
\centering
\setlength{\tabcolsep}{2.8pt}
\begin{tabular}{|c|cccc|cccc|c|}
\hline
\multirow{2}{*}{Model} & \multicolumn{4}{c|}{Actions (6)}                   & \multicolumn{4}{c|}{Locations (*)}                & Average        \\ \cline{2-10} 
                       & Acc            & C   & + C           & - C            & Acc            & C   & + C           & - C           & Acc            \\ \hline
Baseline               & \textbf{73.05} & -   & -             & -              & 68.21          & -   & -             & -             & \textbf{70.47} \\ \hline
Sequential             & 71.56          & 75  & 13.33         & 46.66          & 67.63          & 255 & 27.8          & 32.2          & 69.47          \\ \hline
ILP                    & \textbf{73}             & 63  & \textbf{36.5} & 38.1           & 66.38          & 217 & 19.8          & 35.9          & 69.47          \\
+ Acc                  & \textbf{73}             & 63  & \textbf{36.5} & 38.1           & 66.43          & 217 & 19.8          & 35.9          & 69.50          \\
+ Prior                 & 72.88          & 119 & 31.93         & \textbf{34.45} & 67.54          & 138 & 23.2          & 32.6          & \textbf{70.03}          \\
+ Ent + Acc            & 72.93          & 63  & 34.92         & 38.1           & 66.38          & 219 & 19.6          & 35.6          & 69.44          \\
+ Ent + Prior           & 71.62          & 209 & 25.83         & 37.32          & 68.16          & 53  & 26.4          & 28.3          & 69.78          \\
+ All     & 71.74          & 198 & 25.75         & 36.86          & \textbf{68.27} & 72  & \textbf{29.2} & \textbf{27.8} & 69.89          \\ \hline
\end{tabular}
\caption{Results on Propara dataset. The dataset comprises $1910$ location decisions and $1674$ action decisions. *The output size of location decisions depends on the context of each procedure. 
}
\label{tab:propara}
\end{table}

%% file: tables/new_metrics.tex
\begin{table}[ht]
\footnotesize
\setlength{\tabcolsep}{2.5pt}
\centering
\begin{tabular}{|c|c|c|c|}
\hline
Dataset                        & Model                                                     & Satisfaction & Set Correctness \\ \hline
\multirow{4}{*}{Animal/Flower} & Baseline                                                  & 96.4         & 53.40           \\
                               & Sequential                                                & 100          & \textbf{54.50}           \\
                               & \begin{tabular}[c]{@{}c@{}}ILP\end{tabular} & 100          & \textbf{54.50}          \\
                               & ILP~(Ours)                                         & 100          & \textbf{54.50}           \\ \hline
\multirow{4}{*}{VQA}           & Baseline                                                  & 38.99        & 53.97           \\
                               & Sequential                                                & 100          & 57.66           \\
                               & \begin{tabular}[c]{@{}c@{}}ILP\end{tabular} & 100          & 51.17           \\
                               & ILP~(Ours)                                         & 100          & \textbf{58.27}           \\ \hline
\multirow{4}{*}{Propara}       & Baseline                                                  & 45.12        & 23.30           \\
                               & Sequential                                                & 100          & 28.81           \\
                               & \begin{tabular}[c]{@{}c@{}}ILP\end{tabular} & 100          & 29.9            \\
                               & ILP~(Ours)                                               & 100          & \textbf{30.93}           \\ \hline
\end{tabular}
\caption{Results of our proposed technique, baselines, and expert-written decoding strategies in terms of constraint satisfaction and set correctness. The \textbf{Set Correctness} metric reflects the practical usability of sets of dependent decisions in downstream applications. The new ILP formulation showcased in this table by ILP~(Ours) uses \textit{Entropy + Prior} for the Animal/Flower and VQA task while only utilizing the \textit{Prior} for the Propara task. }
\label{tab:new_metrics}
\end{table}

%% file: appendix.tex
\section{Datasets \& Baselines}
\label{sec:datasets}
\subsection{Animal/Flower/Food}
The dataset\footnote{https://github.com/kaustubh77/Multi-Class-Classification} employed in this study is sourced from the online platform 'Flickr' and encompasses a total of 5439 images classified into three primary categories, namely 'Flower,' 'Animal,' and 'Food.' In the absence of an officially designated test set, a random partitioning strategy is adopted to ensure comparability in the distribution of training and testing instances. Consequently, the resulting splits are utilized within the experimental framework. The training subset encompasses 4531 images, while the test set comprises 1088 images. The dataset further comprises various sub-categories, including 'cat,' 'dog,' 'monkey,' 'squirrel,' 'daisy,' 'dandelion,' 'rose,' 'sunflower,' 'tulip,' 'donuts,' 'lasagna,' 'pancakes,' 'pizza,' 'risotto,' and 'salad.' It should be noted that the data distribution across labels is not balanced, posing a more challenging classification task. This dataset is employed as a simplified scenario to illustrate the benefits of the proposed inference approach.

As the baseline for this task, we use ResNet-50 to represent the images and add a single layer MLP on top for each level. The model is further trained by Cross-Entropy objective and AdamW as optimizer.

The sequential decoding strategy for this dataset propagates labels in a top-down manner, where the highest probable children of the selected Level1 decisions is chosen as the prediction at Level2.

\subsection{20News}
\label{sec:appendix_20news}
This dataset comprises a collection of diverse news articles classified into 23 distinct categories. In order to capture the hierarchical structure inherent in the dataset's labels, we partition these categories into two levels. It should be noted that certain higher-level concepts lack corresponding lower-level labels, necessitating the inclusion of a 'None' label at level 2. Furthermore, we perform a removal process on the initially annotated data containing the 'None' labels, as this subset primarily consists of noisy documents that do not align with any categories present within the dataset. It is crucial to differentiate this removal process from the intentional addition of the 'None' label at level 2, which we manually introduced. 

As the baseline for this task, we initially employed the Bert-Base encoder to generate representations for each news story. Due to the limited context size of Bert, which is constrained to a maximum of 512 tokens, we truncate the news articles accordingly and utilize the CLS token as the representative embedding for the entire article. For Level 1, a 2-layer Multilayer Perceptron (MLP) architecture is employed, with LeakyReLU serving as the chosen activation function. Additionally, Level 2 decisions are made using a single-layer MLP. During the training process, the model is optimized using the AdamW optimizer, with the Cross-Entropy loss function being employed.

The sequential decoding strategy is this dataset is a bottom-up strategy. Here, the model's decision from Level2 is propagated into Level1 without looking further into the initial probabilities generated by the model at that level. 
\subsubsection{Results}
\input{tables/20news}
The baseline performance is similar across different decisions. Thus, considering either the \textit{Expected Accuracy} or the \textit{Prior Probability} in isolation does not have a substantial impact on the global optimization process. However, the inclusion of all proposed factors \textit{(Entropy, Accuracy, and Prior Probability)} leads to a balanced and optimal solution. Although the overall task performance in this experiment does not show significant improvements, this is mainly because the initial decision inconsistencies are minimal. Nevertheless, evaluating the positive and negative changes provides valuable insights into the significance of incorporating the proposed factors.

\subsection{OK-VQA~(COCO)}
The OK-VQA dataset is primarily introduced as a means to propose an innovative task centered around question-answering utilizing external knowledge. To construct this dataset, a subset of the COCO dataset is employed, with augmented annotations obtained through crowdsourcing. While the main objective of the dataset revolves around question answering, it is important to note that it encompasses two levels of annotation. These annotations not only indicate the answer to the given question but also provide additional clarifications regarding the types of objects depicted in the corresponding images. In order to leverage knowledge pertaining to image type relationships, the label set is expanded to include supplementary high-level concepts. Additionally, a knowledge base is provided, delineating parent-child relationships between these labels. The dataset comprises a total of 500 object labels. To enhance the breadth of knowledge encompassed by the dataset, we incorporate additional information from ConceptNet to establish comprehensive relationships among the labels. Notably, both the new information and the original knowledge base may contain noisy information. This, in conjunction with the original knowledge base, forms a four-level hierarchical dependency among the initial 500 labels. Consequently, certain labels within each level may not possess corresponding children at lower levels, necessitating the introduction of 'None' labels at levels 2, 3, and 4.

In this study, we employ the Faster R-CNN framework~\cite{ren2015faster} along with ResNet-110 as the chosen methodology to represent individual objects within images. Subsequently, a one-layer Multilayer Perceptron (MLP) architecture is utilized to classify the images at each level of the hierarchical structure. It should be noted that the number of positive examples (i.e., labels that are not denoted as 'None') decreases as we move toward lower levels of the hierarchy. To address this, we perform subsampling on the 'None' labels for the corresponding classifiers at those levels. The models are trained with the Cross-Entropy loss function and the AdamW optimizer.

The sequential decoding strategy for this dataset is a two-stage top-down and then bottom-up process. Here, `None' labels are first propagated from Level 1 to Level 4, and then the selected label~(if not None) from Level 4 is propagated bottom-up to Level 1. Since each label at level$n$ only has one parent in Level$n-1$, this process does not need to look into the original model probabilities for propagation. 

\subsection{Propara}
\begin{figure}[h]
    \centering
    \includegraphics[width=\linewidth]{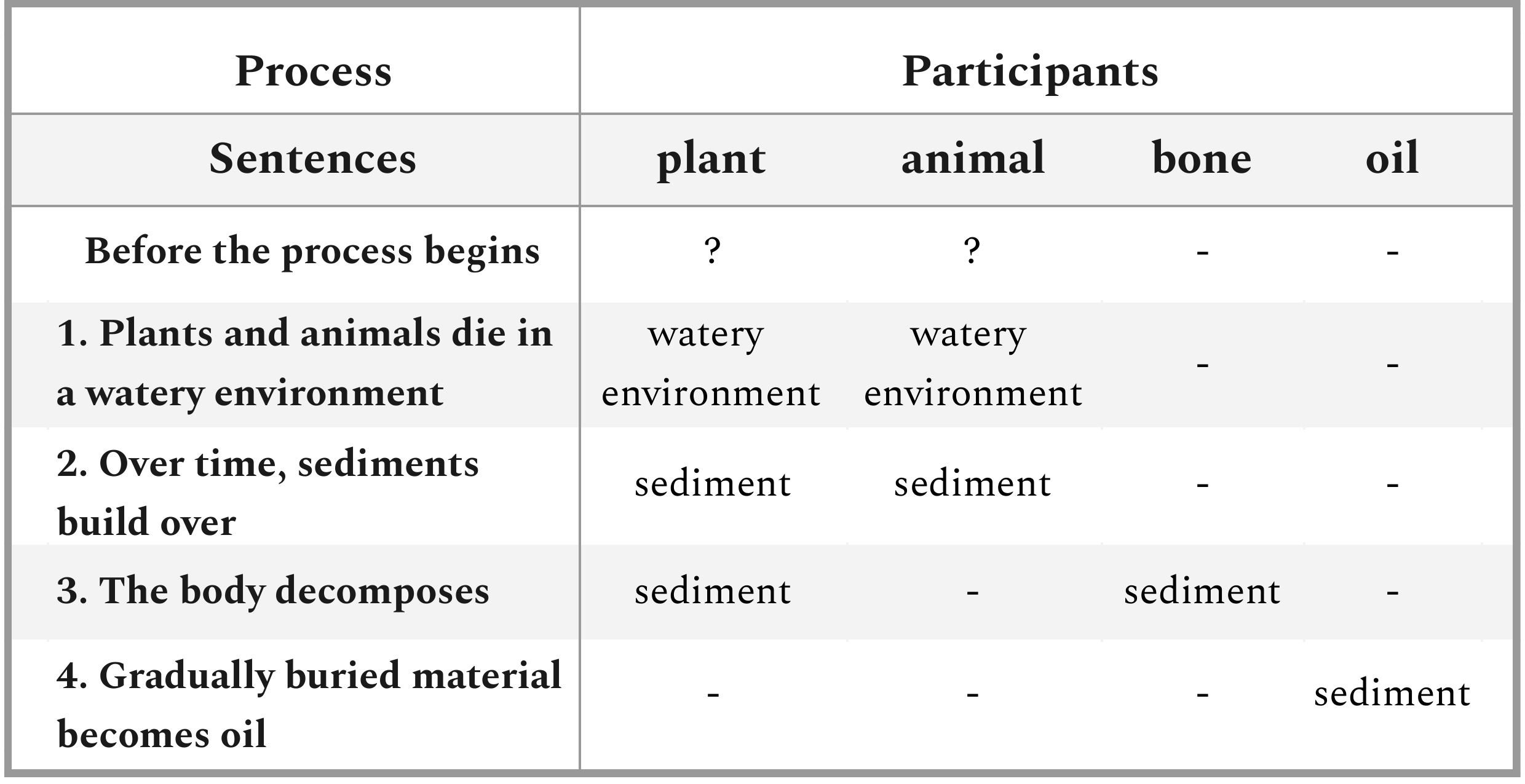}
    \caption{An example from the Propara dataset taken from \cite{faghihi2023role}. `-' refers to the entity not existing; `?' refers to the entity whose location is unclear.
    }
    \label{fig:propara}
\end{figure}

The Propara dataset serves as a procedural reasoning benchmark, primarily devised to assess the ability of models to effectively track significant entities across a series of events. The stories within this dataset revolve around natural phenomena, such as photosynthesis. The annotation process involves capturing crucial entities and their corresponding locations at each step of the process, which are obtained through crowd-sourcing efforts. An illustrative example of this dataset is depicted in Figure \ref{fig:propara}.

The sequence of locations pertaining to each entity can be further extended to infer the actions or status of the entity at each step. Previous studies~\cite{dalvi2019everything} have proposed six possible actions for each entity at each step, namely 'Create,' 'Move,' 'Exist,' 'Destroy,' 'Prior,' and 'Post.' In this context, 'Prior' signifies an entity that has not yet been created, while 'Post' denotes an entity that has already been destroyed. 

As for the baseline, we employ a modified version of the MeeT~\cite{singh-etal-2023-entity} architecture. The architecture utilizes T5-Large~\cite{raffel2020exploring} as the backbone and employs a Question-Answering framework to extract the location and action of each entity at each step. The format of the input to the model is as follows for entity $e$ and step $i$: "Where is $e$ located in sent $i$? Sent 1: ..., Sent 2: ..., ...". For extracting the action, the set of options is also passed as input, resulting in the modification of the question to "What is the status of entity $e$ in sent $i$? (a) Create (b) Move (c) Destroy (d) Exist (e) Prior (f) Post". 

Although the original model of MeeT incorporates a Conditional Random Field~(CRF)~\cite{lafferty2001conditional} layer during inference to ensure consistency among action decisions, we exclude this layer from our baseline. This decision is motivated by two reasons. Firstly, the use of CRF in this context is not generalizable as it relies on training data statistics for defining transitional scores. Secondly, we intend to impose consistency using various inference mechanisms on our end and consider a joint framework to ensure both locations and actions exhibit consistency. Additionally, while the MeeT baseline employs two independent T5-Large models for each question type (location and action), our baseline utilizes the same model for both question types.
For the sequential decoding technique to enforce sequential consistency among the series of interrelated action and location decisions, we utilize the post-processing code presented in \citet{faghihi2023role}.

\section{Metrics}
\label{sec:metrics_appendix}
Here, we briefly describe the metrics used in this paper to evaluate the methods.
\subsection{Number of Changes}
This metric quantifies the post-inference changes in decisions, specifically assessing the extent to which original decisions are altered due to inference constraints. It serves as a crucial indicator of whether the optimization method treats all decisions equally or exhibits a preference for certain decisions over others. A genuinely global optimization method will result in multiple decision changes, promoting a more balanced distribution of alterations across all decisions. In contrast, expert-written strategies tend to favor specific decisions. This metric is straightforward to calculate by comparing the differences between decisions before and after applying the inference mechanism.
\subsection{Ratio of In-Correct to Correct Changes~(+C)}
This metric reveals the proportion of post-inference changes that are deemed favorable. While this metric may not carry substantial standalone significance, it serves as a valuable means to compare different inference techniques. A higher ratio signifies that the inference method has been more successful in deducing accurate labels based on the imposed constraints.

\subsection{Ratio of Correct to In-Correct Changes~(-C)}
This number shows the extent of undesirable changes made after inference. A lower ratio means the inference method has done a better job of preventing errors while ensuring the output adheres to the constraints.

\subsection{Satisfaction Rate}
This metric shows how well predictions align with constraints. We calculate it by generating constraint instances from related decisions and counting the satisfying cases against all possible instances. Inference techniques guarantee that modified decisions always adhere to the constraints, resulting in a satisfaction rate of 100\%.

\subsection{Correctly Predicated Sets of Interrelated Decisions}
This metric is crucial for assessing the practical usefulness of the output from inference techniques or the original network decisions in downstream applications. The primary objective of inference mechanisms is to boost the percentage of these fully satisfying cases compared to the model's original performance, all while ensuring that the decisions align with the task's constraints. For instance, in a hierarchical classification task, we consider one instance to be correct only when the decisions at all levels are simultaneously accurate.

\section{Discussion}
Here, we address some of the key questions about this work. 

\subsection{Q1: Which metric is most important among the ones evaluated in this paper?}
All the metrics assessed in this paper provide insights into the model's performance. Among these, the \textbf{Set Correctness} score offers a comprehensive evaluation that combines constraint satisfaction and correctness, indicating the proportion of output decisions suitable for safe use in downstream tasks.

When comparing different ILP variations, the primary focus should be on the original task performance since they all share the same high satisfaction score of 100\%. Additionally, the \textbf{Change} metric helps reveal whether an ILP variation conducts truly global optimization or exhibits a bias towards specific prediction classes.

In the context of comparing the baseline method with inference techniques, it is essential to consider both the \textbf{satisfaction} and \textbf{set correctness} scores. This is because the raw model predictions, as initially generated, may not be directly acceptable. For instance, if a model predicts a ``Move'' action for entity A at step 4, but the location prediction does not indicate a change in location, it becomes unclear whether entity A indeed changed locations or not. 

\subsection{Why utilize the model's overall accuracy in the score function instead of its accuracy for a specific decision variable?}

In our context, we assume that each decision type corresponds to a specific model. Therefore, assessing the model's accuracy is the same as evaluating the accuracy of a particular decision type. If a single model supplies multiple decision types, we can easily expand this concept to evaluate the accuracy of each decision type individually within the same framework. 

\subsection{What is the main difference between the sequential decoding strategy and the ILP formulation?}

The sequential decoding strategy is a domain-specific, expert-crafted technique employed for addressing decision inconsistencies in accordance with task constraints. In contrast, the ILP (Integer Linear Programming) formulation offers a more general, non-customized approach that isn't tailored to individual tasks.

Sequential decoding strategies typically involve rules or programs that often exhibit a preference for a specific decision while adjusting other decisions to align with it. This approach tends to prioritize decision alignment over considering the probabilities associated with these decisions. On the other hand, the ILP optimization process seeks the most optimized solution by taking into account the raw probabilities from the models and the imposed constraints.

%% file: tables/20news.tex
\begin{table}[t]
\scriptsize
\centering
\setlength{\tabcolsep}{3pt}
\begin{tabular}{|c|cccc|ccc|c|}
\hline
\multirow{2}{*}{Model} & \multicolumn{4}{c|}{Level 1 (16)}                               & \multicolumn{3}{c|}{Level 2 (8)}              & Average        \\ \cline{2-9} 
                       & F1             & C & + C      & - C      & F1             & C & + C      & F1             \\ \hline
Baseline               & 73.62          & -       & -              & -              & 75.13          & -       & -              & 74.01          \\ \hline
Sequential             & 72.99          & 330     & 20.6           & 46.36              & 75.13          & 0       & 0.00           & 73.55          \\ \hline
ILP                    & 73.53          & 225     & 25.78          & 39.55          & 75.46          & 68      & 63.24          & 74.03          \\
+ Acc                  & 73.57          & 212     & \textbf{26.89}          & 39.62          & 75.45          & 73      & 64.39          & 74.05          \\
+ Prior                 & 73.35          & 161     & 25.46          & 39.13          & 75.35          & 94      & 65.96          & 74.01          \\
+ Ent + Acc            & 73.54          & 205     & 26.34 & 40             & 75.39          & 75      & \textbf{64}             & 74.02          \\
+ Ent + Prior           & 73.63          & 125     & 26.4           & 36             & 75.49          & 112     & 68.75 & 74.12          \\
+ All     & \textbf{73.64} & 131     & 25.95          & \textbf{35.11} & \textbf{75.52} & 111     & 68.47          & \textbf{74.13} \\ \hline
\end{tabular}
\caption{Results on 20News dataset. Here, the \textit{-C} of level 2 is $0$ in all cases.}
\label{tab:20news}
\end{table}

%% file: main.bbl
\begin{thebibliography}{40}
\expandafter\ifx\csname natexlab\endcsname\relax\def\natexlab#1{#1}\fi

\bibitem[{Anderson et~al.(2017)Anderson, Fernando, Johnson, and
  Gould}]{anderson2017guided}
Peter Anderson, Basura Fernando, Mark Johnson, and Stephen Gould. 2017.
\newblock Guided open vocabulary image captioning with constrained beam search.
\newblock In \emph{Proceedings of the 2017 Conference on Empirical Methods in
  Natural Language Processing}, pages 936--945.

\bibitem[{Bosselut et~al.(2018)Bosselut, Levy, Holtzman, Ennis, Fox, and
  Choi}]{bosselut2017simulating}
Antoine Bosselut, Omer Levy, Ari Holtzman, Corin Ennis, Dieter Fox, and Yejin
  Choi. 2018.
\newblock Simulating action dynamics with neural process networks.
\newblock In \emph{Proceedings of the 6th International Conference for Learning
  Representations (ICLR)}.

\bibitem[{Chang et~al.(2012)Chang, Ratinov, and Roth}]{chang2012structured}
Ming-Wei Chang, Lev Ratinov, and Dan Roth. 2012.
\newblock Structured learning with constrained conditional models.
\newblock \emph{Machine learning}, 88(3):399--431.

\bibitem[{Dahlmeier and Ng(2012)}]{dahlmeier2012beam}
Daniel Dahlmeier and Hwee~Tou Ng. 2012.
\newblock A beam-search decoder for grammatical error correction.
\newblock In \emph{EMNLP 2012}, pages 568--578.

\bibitem[{Dalvi et~al.(2018)Dalvi, Huang, Tandon, Yih, and
  Clark}]{dalvi2018tracking}
Bhavana Dalvi, Lifu Huang, Niket Tandon, Wen-tau Yih, and Peter Clark. 2018.
\newblock Tracking state changes in procedural text: a challenge dataset and
  models for process paragraph comprehension.
\newblock In \emph{Proceedings of the 2018 Conference of the North American
  Chapter of the Association for Computational Linguistics: Human Language
  Technologies, Volume 1 (Long Papers)}, pages 1595--1604.

\bibitem[{Dalvi et~al.(2019)Dalvi, Tandon, Bosselut, Yih, and
  Clark}]{dalvi2019everything}
Bhavana Dalvi, Niket Tandon, Antoine Bosselut, Wen-tau Yih, and Peter Clark.
  2019.
\newblock Everything happens for a reason: Discovering the purpose of actions
  in procedural text.
\newblock In \emph{(EMNLP-IJCNLP)}, pages 4496--4505.

\bibitem[{Devlin et~al.(2019)Devlin, Chang, Lee, and
  Toutanova}]{kenton2019bert}
Jacob Devlin, Ming-Wei Chang, Kenton Lee, and Kristina Toutanova. 2019.
\newblock \href {https://doi.org/10.18653/v1/N19-1423} {{BERT}: Pre-training of
  deep bidirectional transformers for language understanding}.
\newblock In \emph{Proceedings of the 2019 Conference of the North {A}merican
  Chapter of the Association for Computational Linguistics: Human Language
  Technologies, Volume 1 (Long and Short Papers)}, pages 4171--4186,
  Minneapolis, Minnesota. Association for Computational Linguistics.

\bibitem[{Faghihi et~al.(2021)Faghihi, Guo, Uszok, Nafar, and
  Kordjamshidi}]{faghihi2021domiknows}
Hossein~Rajaby Faghihi, Quan Guo, Andrzej Uszok, Aliakbar Nafar, and Parisa
  Kordjamshidi. 2021.
\newblock Domiknows: A library for integration of symbolic domain knowledge in
  deep learning.
\newblock In \emph{EMNLP: System Demonstrations}, pages 231--241.

\bibitem[{Faghihi and Kordjamshidi(2021)}]{faghihi2021timestamped}
Hossein~Rajaby Faghihi and Parisa Kordjamshidi. 2021.
\newblock Time-stamped language model: Teaching language models to understand
  the flow of events.
\newblock In \emph{Proceedings of the 2021 Conference of the North American
  Chapter of the Association for Computational Linguistics: Human Language
  Technologies}, pages 4560--4570.

\bibitem[{Faghihi et~al.(2023{\natexlab{a}})Faghihi, Kordjamshidi, Teng, and
  Allen}]{faghihi2023role}
Hossein~Rajaby Faghihi, Parisa Kordjamshidi, Choh~Man Teng, and James Allen.
  2023{\natexlab{a}}.
\newblock The role of semantic parsing in understanding procedural text.
\newblock In \emph{Findings of the Association for Computational Linguistics:
  EACL 2023}, pages 1792--1804.

\bibitem[{Faghihi et~al.(2023{\natexlab{b}})Faghihi, Nafar, Zheng, Mirzaee,
  Zhang, Uszok, Wan, Premsri, Roth, and Kordjamshidi}]{faghihi2023gluecons}
Hossein~Rajaby Faghihi, Aliakbar Nafar, Chen Zheng, Roshanak Mirzaee, Yue
  Zhang, Andrzej Uszok, Alexander Wan, Tanawan Premsri, Dan Roth, and Parisa
  Kordjamshidi. 2023{\natexlab{b}}.
\newblock Gluecons: A generic benchmark for learning under constraints.
\newblock \emph{arXiv preprint arXiv:2302.10914}.

\bibitem[{Freitag and Al-Onaizan(2017)}]{freitag2017beam}
Markus Freitag and Yaser Al-Onaizan. 2017.
\newblock Beam search strategies for neural machine translation.
\newblock \emph{ACL 2017}, page~56.

\bibitem[{Gal and Ghahramani(2016)}]{gal2016dropout}
Yarin Gal and Zoubin Ghahramani. 2016.
\newblock Dropout as a bayesian approximation: Representing model uncertainty
  in deep learning.
\newblock In \emph{international conference on machine learning}, pages
  1050--1059. PMLR.

\bibitem[{Guo et~al.(2021)Guo, Faghihi, Zhang, Uszok, and
  Kordjamshidi}]{guo2021inference}
Quan Guo, Hossein~Rajaby Faghihi, Yue Zhang, Andrzej Uszok, and Parisa
  Kordjamshidi. 2021.
\newblock Inference-masked loss for deep structured output learning.
\newblock In \emph{Proceedings of the Twenty-Ninth International Conference on
  International Joint Conferences on Artificial Intelligence}, pages
  2754--2761.

\bibitem[{Guo et~al.(2020)Guo, Rajaby~Faghihi, Zhang, Uszok, and
  Kordjamshidi}]{inference-ijcai2020-382}
Quan Guo, Hossein Rajaby~Faghihi, Yue Zhang, Andrzej Uszok, and Parisa
  Kordjamshidi. 2020.
\newblock \href {https://doi.org/10.24963/ijcai.2020/382} {Inference-masked
  loss for deep structured output learning}.
\newblock In \emph{Proceedings of the Twenty-Ninth International Joint
  Conference on Artificial Intelligence, {IJCAI-20}}, pages 2754--2761.
  International Joint Conferences on Artificial Intelligence Organization.
\newblock Main track.

\bibitem[{{Gurobi Optimization, LLC}(2023)}]{gurobi}
{Gurobi Optimization, LLC}. 2023.
\newblock \href {https://www.gurobi.com} {{Gurobi Optimizer Reference Manual}}.

\bibitem[{He et~al.(2016)He, Zhang, Ren, and Sun}]{he2016deep}
Kaiming He, Xiangyu Zhang, Shaoqing Ren, and Jian Sun. 2016.
\newblock Deep residual learning for image recognition.
\newblock In \emph{Proceedings of the IEEE conference on computer vision and
  pattern recognition}, pages 770--778.

\bibitem[{Hu et~al.(2016)Hu, Ma, Liu, Hovy, and Xing}]{hu2016harnessing}
Zhiting Hu, Xuezhe Ma, Zhengzhong Liu, Eduard Hovy, and Eric Xing. 2016.
\newblock Harnessing deep neural networks with logic rules.
\newblock In \emph{54th ACL}, pages 2410--2420.

\bibitem[{Jiang et~al.(2023)Jiang, Ilievski, and Ma}]{jiang2023transferring}
Yifan Jiang, Filip Ilievski, and Kaixin Ma. 2023.
\newblock Transferring procedural knowledge across commonsense tasks.
\newblock \emph{arXiv preprint arXiv:2304.13867}.

\bibitem[{Kordjamshidi and Moens(2015)}]{SemanticWeb}
Parisa Kordjamshidi and Marie-Francine Moens. 2015.
\newblock \href {https://doi.org/10.1016/j.websem.2014.06.001} {Global machine
  learning for spatial ontology population}.
\newblock \emph{Web Semant.}, 30(C):3--21.

\bibitem[{Lafferty et~al.(2001)Lafferty, McCallum, and
  Pereira}]{lafferty2001conditional}
John Lafferty, Andrew McCallum, and Fernando~CN Pereira. 2001.
\newblock Conditional random fields: Probabilistic models for segmenting and
  labeling sequence data.

\bibitem[{Lin et~al.(2014)Lin, Maire, Belongie, Hays, Perona, Ramanan,
  Doll{\'a}r, and Zitnick}]{lin2014microsoft}
Tsung-Yi Lin, Michael Maire, Serge Belongie, James Hays, Pietro Perona, Deva
  Ramanan, Piotr Doll{\'a}r, and C~Lawrence Zitnick. 2014.
\newblock Microsoft coco: Common objects in context.
\newblock In \emph{Computer Vision--ECCV 2014: 13th European Conference,
  Zurich, Switzerland, September 6-12, 2014, Proceedings, Part V 13}, pages
  740--755. Springer.

\bibitem[{Liu et~al.(2022)Liu, Jiang, Monath, Cotterell, and
  Sachan}]{liu2022autoregressive}
Tianyu Liu, Yuchen Jiang, Nicholas Monath, Ryan Cotterell, and Mrinmaya Sachan.
  2022.
\newblock Autoregressive structured prediction with language models.
\newblock \emph{arXiv preprint arXiv:2210.14698}.

\bibitem[{Lu et~al.(2021)Lu, West, Zellers, Le~Bras, Bhagavatula, and
  Choi}]{lu2021neurologic}
Ximing Lu, Peter West, Rowan Zellers, Ronan Le~Bras, Chandra Bhagavatula, and
  Yejin Choi. 2021.
\newblock Neurologic decoding:(un) supervised neural text generation with
  predicate logic constraints.
\newblock In \emph{Proceedings of the 2021 Conference of the North American
  Chapter of the Association for Computational Linguistics: Human Language
  Technologies}, pages 4288--4299.

\bibitem[{Marino et~al.(2019)Marino, Rastegari, Farhadi, and
  Mottaghi}]{marino2019ok}
Kenneth Marino, Mohammad Rastegari, Ali Farhadi, and Roozbeh Mottaghi. 2019.
\newblock Ok-vqa: A visual question answering benchmark requiring external
  knowledge.
\newblock In \emph{Proceedings of the IEEE/cvf conference on computer vision
  and pattern recognition}, pages 3195--3204.

\bibitem[{Nandwani et~al.(2019)Nandwani, Pathak, Singla
  et~al.}]{nandwani2019primal}
Yatin Nandwani, Abhishek Pathak, Parag Singla, et~al. 2019.
\newblock A primal dual formulation for deep learning with constraints.
\newblock In \emph{Advances in Neural Information Processing Systems}, pages
  12157--12168.

\bibitem[{Ning et~al.(2018)Ning, Feng, Wu, and Roth}]{ning2018joint}
Qiang Ning, Zhili Feng, Hao Wu, and Dan Roth. 2018.
\newblock Joint reasoning for temporal and causal relations.
\newblock In \emph{Proceedings of the 56th Annual Meeting of the Association
  for Computational Linguistics (Volume 1: Long Papers)}, pages 2278--2288.

\bibitem[{Punyakanok et~al.(2004)Punyakanok, Roth, Yih, and
  Zimak}]{punyakanok2004semantic}
Vasin Punyakanok, Dan Roth, Wen-tau Yih, and Dav Zimak. 2004.
\newblock Semantic role labeling via integer linear programming inference.
\newblock In \emph{COLING 2004: Proceedings of the 20th International
  Conference on Computational Linguistics}, pages 1346--1352.

\bibitem[{Raffel et~al.(2020)Raffel, Shazeer, Roberts, Lee, Narang, Matena,
  Zhou, Li, and Liu}]{raffel2020exploring}
Colin Raffel, Noam Shazeer, Adam Roberts, Katherine Lee, Sharan Narang, Michael
  Matena, Yanqi Zhou, Wei Li, and Peter~J Liu. 2020.
\newblock Exploring the limits of transfer learning with a unified text-to-text
  transformer.
\newblock \emph{Journal of Machine Learning Research}, 21:1--67.

\bibitem[{Rajaby~Faghihi et~al.(2021)Rajaby~Faghihi, Guo, Uszok, Nafar, and
  Kordjamshidi}]{rajaby-faghihi-etal-2021-domiknows}
Hossein Rajaby~Faghihi, Quan Guo, Andrzej Uszok, Aliakbar Nafar, and Parisa
  Kordjamshidi. 2021.
\newblock \href {https://doi.org/10.18653/v1/2021.emnlp-demo.27}
  {{D}omi{K}now{S}: A library for integration of symbolic domain knowledge in
  deep learning}.
\newblock In \emph{Proceedings of the 2021 Conference on Empirical Methods in
  Natural Language Processing: System Demonstrations}, pages 231--241, Online
  and Punta Cana, Dominican Republic. Association for Computational
  Linguistics.

\bibitem[{Ren et~al.(2015)Ren, He, Girshick, and Sun}]{ren2015faster}
Shaoqing Ren, Kaiming He, Ross Girshick, and Jian Sun. 2015.
\newblock Faster r-cnn: Towards real-time object detection with region proposal
  networks.
\newblock \emph{Advances in neural information processing systems}, 28.

\bibitem[{Rizzolo and Roth(2016)}]{rizzolo2016integer}
Nick Rizzolo and Dan Roth. 2016.
\newblock Integer linear programming for coreference resolution.
\newblock \emph{Anaphora Resolution: Algorithms, Resources, and Applications},
  pages 315--343.

\bibitem[{Roth and Yih(2005)}]{roth2005integer}
Dan Roth and Wen-tau Yih. 2005.
\newblock Integer linear programming inference for conditional random fields.
\newblock In \emph{Proceedings of the 22nd international conference on Machine
  learning}, pages 736--743.

\bibitem[{Scholak et~al.(2021)Scholak, Schucher, and
  Bahdanau}]{scholak2021picard}
Torsten Scholak, Nathan Schucher, and Dzmitry Bahdanau. 2021.
\newblock Picard: Parsing incrementally for constrained auto-regressive
  decoding from language models.
\newblock In \emph{EMNLP}, pages 9895--9901.

\bibitem[{Singh et~al.(2023)Singh, Bai, and Wang}]{singh-etal-2023-entity}
Janvijay Singh, Fan Bai, and Zhen Wang. 2023.
\newblock \href {https://aclanthology.org/2023.eacl-main.90} {Entity tracking
  via effective use of multi-task learning model and mention-guided decoding}.
\newblock In \emph{Proceedings of the 17th Conference of the European Chapter
  of the Association for Computational Linguistics}, pages 1255--1263,
  Dubrovnik, Croatia. Association for Computational Linguistics.

\bibitem[{Wang et~al.(2022)Wang, Liu, Chen, Hong, Tang, and
  Song}]{wang2022deepstruct}
Chenguang Wang, Xiao Liu, Zui Chen, Haoyun Hong, Jie Tang, and Dawn Song. 2022.
\newblock Deepstruct: Pretraining of language models for structure prediction.
\newblock In \emph{Findings of the Association for Computational Linguistics:
  ACL 2022}, pages 803--823.

\bibitem[{Xiao and Wang(2019)}]{xiao2019quantifying}
Yijun Xiao and William~Yang Wang. 2019.
\newblock Quantifying uncertainties in natural language processing tasks.
\newblock In \emph{Proceedings of the AAAI conference on artificial
  intelligence}, volume~33, pages 7322--7329.

\bibitem[{Xu et~al.(2018)Xu, Zhang, Friedman, Liang, and
  Broeck}]{xu2018semantic}
Jingyi Xu, Zilu Zhang, Tal Friedman, Yitao Liang, and Guy Broeck. 2018.
\newblock A semantic loss function for deep learning with symbolic knowledge.
\newblock In \emph{ICML}, pages 5502--5511. PMLR.

\bibitem[{Young et~al.(2014)Young, Lai, Hodosh, and
  Hockenmaier}]{young2014image}
Peter Young, Alice Lai, Micah Hodosh, and Julia Hockenmaier. 2014.
\newblock From image descriptions to visual denotations: New similarity metrics
  for semantic inference over event descriptions.
\newblock \emph{Transactions of the Association for Computational Linguistics},
  2:67--78.

\bibitem[{Zhu and Laptev(2017)}]{zhu2017deep}
Lingxue Zhu and Nikolay Laptev. 2017.
\newblock Deep and confident prediction for time series at uber.
\newblock In \emph{2017 IEEE International Conference on Data Mining Workshops
  (ICDMW)}, pages 103--110. IEEE.

\end{thebibliography}
